# Improving Chronic Kidney Disease Detection Efficiency: Fine Tuned CatBoost and Nature-Inspired Algorithms with Explainable AI

Md. Ehsanul Haque
Department of Computer Science and Engineering
East West University
Dhaka, Bangladesh
ehsanulhaquesohan758@gmail.com

S. M. Jahidul Islam
Department of Computer Science and Mathematics
Bangladesh Agricultural University
Mymensingh, Bangladesh
jahidul.ict@bau.edu.bd

Jeba Maliha
Department of Computer Science and Engineering
Ahsanullah University of Science and Technology
Dhaka, Bangladesh
malihajebasqa@gmail.com

Md. Shakhauat Hossan Sumon
Department of Electrical and Computer Engineering
North South University
Dhaka, Bangladesh
sumon.eee.cse@gmail.com

Rumana Sharmin
Department of Food and Nutrition
University of Dhaka
Dhaka, Bangladesh
rumanasharminshorna@gmail.com

Sakib Rokoni
Department of Computer Science and Engineering
BRAC University
Dhaka, Bangladesh
sakibrokoni61@gmail.com

*Abstract*—Chronic Kidney Disease (CKD) is a major global health issue which is affecting million people around the world and with increasing rate of mortality. Mitigation of progression of CKD and better patient outcomes requires early detection. Nevertheless, limitations lie in traditional diagnostic methods, especially in resource constrained settings. This study proposes an advanced machine learning approach to enhance CKD detection by evaluating four models: Random Forest (RF), Multi-Layer Perceptron (MLP), Logistic Regression (LR), and a fine-tuned CatBoost algorithm. Specifically, among these, the fine-tuned CatBoost model demonstrated the best overall performance having an accuracy of 98.75%, an AUC of 0.9993 and a Kappa score of 97.35% of the studies. The proposed CatBoost model has used a nature inspired algorithm such as Simulated Annealing to select the most important features, Cuckoo Search to adjust outliers and grid search to fine tune its settings in such a way to achieve improved prediction accuracy. Features significance is explained by SHAP-a well-known XAI technique-for gaining transparency in the decision-making process of proposed model and bring up trust in diagnostic systems. Using SHAP, the significant clinical features were identified as specific gravity, serum creatinine, albumin, hemoglobin, and diabetes mellitus. The potential of advanced machine learning techniques in CKD detection is shown in this research, particularly for low income and middle-income healthcare settings where prompt and correct diagnoses are vital. This study seeks to provide a highly accurate, interpretable, and efficient diagnostic tool to add to efforts for early intervention and improved healthcare outcomes for all CKD patients.

*Index Terms*—Chronic Kidney Disease, Machine Learning, Nature-Inspired Algorithms, Explainable AI, CKD Detection.

## I. INTRODUCTION

Chronic Kidney Disease (CKD) is a global health problem on the rise and is now among the top causes of death. An estimated 5 to 11 million deaths per year are the global mortality from all causes of kidney disease [1]. By 2040, CKD is expected to be the fifth leading cause of years lived lost [2]. For people in low- and middle-income countries (LMICs) and low-income countries (LICs), limited access to healthcare, late diagnosis and cash challenges take a heavy toll [3]. In Bangladesh, for instance, 22.48% of persons have CKD, and 80% of death cases related to kidney are 'financial barriers' affected [4] [5]. Also, in India, CKD affects 13–15.04% of the population with a worsening of rates of diabetes and hypertension [6].

Although CKD is detected better in high income countries such as the United States, with better public awareness and treatment options, delays in its detection still result in worse health. These delays are driven by a lack of awareness, education, mild symptoms, poor screening, delayed follow up and diagnostic bias [7]. Here also, healthcare access and fragmented care systems coupled with the former also contribute to the problem in LMICs and LICs.

Machine learning (ML) offers a potential solution to improve CKD detection by analyzing complex data to make more accurate predictions. Nevertheless, the problems of irrelevant features, high false positives, false negatives, and lack of understanding of model decisions persist. In addition, the results in the data can be skewed by outliers, and can decrease the performance of the model. To handle these challenges, this paper employs Simulated Annealing (SA) to choose the most discriminative and important features to detect CKD. Moreover, to improve the robustness of the model, outlier adjustment by Cuckoo Search is employed to make the model robust ensuring that the model remains accurate and reliable when outliers are present. These combined techniques result in lower errors and better overall performance of the model in the detection of CKD.

The objectives of this study are:

- Use of Machine learning in enhancing accuracy of CKD detection.
- Cuckoo Search algorithm is used to access the optimization of the outlier adjustment.
- Feature selection to determine the most important variables is done using Simulated Annealing (SA).
- To improve detection, fine tuned CatBoost is proposed.
- Reduce the rate of false positive and false negative.
- Employ XAI methods in order to analyze the decision making process of the model.

This paper intends to use powerful and state-of-the-art machine learning algorithms coupled with reliable feature selection and explainability to reliably develop a technique that can accurately and efficiently diagnose CKD at its early stage. The findings can potentially facilitate earlier diagnosis, enhance prognosis and may help millions of people worldwide especially in LMICs.

## II. LITERATURE REVIEW

In this section we review studies related to detection of Chronic Kidney Disease and their method, strengths, weakness and results for better understanding of what has been done and how it is working.

Shahid et al. [8] mention how Boosting Algorithms such as Gradient Boosting and XGBoost were effective for CKD prediction and the necessity of Preprocessing and Feature Selection.Results show that AdaBoost with 100% accuracy on training set and 98.47% on test set with nearly perfect precision, recall, AUC-ROC. This work offers a strength of ensemble learning in improving accuracy, and provides comprehensive analysis using robust evaluation metrics. limitations are lack of external validation, computational cost and overfitting.

To diagnose Chronic Kidney Disease (CKD), Gangani et al. [9] proposed a machine learning framework with explainable AI (XAI).). The top performing model turned out to be Extreme Gradient Boosting (XGB), which has 100% training and 97.5% testing accuracy. The interpretability was done using SHAP and PDP techniques and the most important features were discovered as specific gravity and hemoglobin. Strengths included high accuracy, effective XAI integration, and user friendly interface, but also limited to potential overfitting, and external validation.

In [10], Saurabh Pal used machine learning to predict Chronic Kidney Disease (CKD) from a UCI data set with categorical and non categorical attribute. On different data combinations, models were evaluated on after preprocessing, and the highest accuracy 93% on categorical attributes was found with Random Forest.The study underscored the importance of early CKD detection but faced limitations, including potential over fitting , and the absence of explainability.

Starting with 25 variables and filtering with predictive modeling, Ariful et al. [11] looked at machine learning approaches for early CKD detection. An evaluation of twelve methods showed the potential of advanced analytics including predictive modeling that improved upon feature selection and efficiency achieving a highest accuracy of 98.3%. However, the immediate applicability in real world healthcare settings is limited by its absence of a thorough comparison with clinical diagnostic standards.

Nayeem et al. [12] used machine learning to predict chronic kidney disease with ANN and reached the highest accuracy of 98.61 % . Preprocessing techniques were shown to be very strong such as imputation and scaling in order to achieve good model performance. Although it did not have detailed class imbalance handling and external validation, its applicability to real world scenarios is limited.

A dataset from the UCI repository with 25 features is used for chronic kidney disease (CKD) by machine learning based techniques in [13] to develop a predictive model. Logistic Regression (LR), Decision Tree (DT) and Support Vector Machine (SVM) classifiers were used in the study and later replaced by a bagging ensemble method to improve model performance. Standalone accuracy of 95.92% was obtained by the Decision Tree model. When the bagging ensemble approach is applied, accuracy increased to 97.23%. Robustness and predictive accuracy of the model were highly dependent on the ensemble methods used.

The J48 algorithm was used to predict CKD stage using machine learning by Ilyas et al. [14] with a dataset from the UCI repository. This study first did preprocessing for the attributes such as age, sex, race, and serum creatinine and then estimated GFR using the CKD-EPI equation. This is validated with 15 fold cross validation, and J48 attained accuracy of 85.5% with execution time of 0.03 seconds, proving it to be efficient and reliable for CKD stage prediction. Nevertheless, use of limited features indicates the potential need of further research to improve robustness and generalizability.

In this work Chaity et al. [15] predicted Chronic Kidney Disease (CKD) using a dataset from the UCI Machine Learning Repository using deep learning algorithms. Imputing missing values and scaling nominal variables was required in preprocessing. Optimized CNN, ANN, and LSTM models were implemented in the study, and the Optmimized CNN model exhibited highest accuracy of 98.75%. Precision, recall, F1 score and AUC were the evaluation metric. Although the study delivered excellent model performance with good results, there was no cross validation and overfitting in the deep learning models was observed.

## III. METHODOLOGY

### A. Data Collection

This study used the dataset from the UCI Machine Learning Repository [16]. It has 400 rows and 26 columns altogether. There are 14 numerical columns and 12 categorical columns out of these. Chronic Kidney Disease (CKD) contains 250 samples labeled as CKD and 150 as non-CKD, as the target variable. The data description of the all features is given below in table I.

TABLE I
DATASET DESCRIPTION

| Variable Name | Description |
|---|---|
| age, bp | Patient's age, blood pressure |
| sg, al, su | Specific gravity, albumin, sugar (urine) |
| rbc, pc, pcc | Red blood cells, pus cells, pus clumps |
| ba, bgr, bu | Bacteria (urine), blood glucose, blood urea |
| sc, sod, pot | Serum creatinine, sodium, potassium |
| hemo, pcv | Hemoglobin, packed cell volume |
| wbcc, rbcc | White and red blood cell count |
| htn, dm, cad | Hypertension, diabetes, coronary artery disease |
| appet, pe, ane | Appetite, pedal edema, anemia |
| class | CKD status (target) |

### B. Preprocessing

- **Encoding:** Categorical columns are encoded by label encoding, and each of the 14 categorical columns is converted to a numeric value. This approach assigns to each category a unique integer while preserving the order of the category.
- **Handle Missing Values:** KNN imputation [17] is used to fill in missing values, inferring missing data using nearest neighbors. For categorical variables, we use mode imputation: that is, replacing missing values with the most frequent value. The distribution of missing values for each column is visualized in Figure 1.

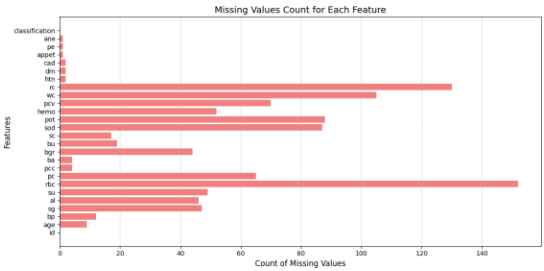

Fig. 1. Missing Values Count for Each Feature

- **Outlier Adjustment:** Cuckoo Search [18] Algorithm inspired by nature mimics cuckoo birds laying eggs in other nest and detects, and corrects the outliers. It iteratively optimizes the dataset and replaces the outliers with a reasonable value in order to make the analysis consistent. This process of outlier adjustment is outlined in Algorithm 1.

**Algorithm 1** Cuckoo Search Algorithm for Outlier Adjustment

**Input:** Initial population of nests
**Output:** Best solution (adjusted dataset)
Set best_nest = ∞ **for** iteration = 1 **to** max_iter **do**
   **for** each nest in population **do**
      Evaluate fitness of the nest **if** fitness is better than best_nest **then**
         Update best_nest
   Generate new nests by applying random perturbation (cuckoo behavior) **for** each new nest **do**
      Evaluate fitness of the new nest **if** fitness is better than best_nest **then**
         Update best_nest
   Replace old population with new nests
**return** best_nest (adjusted dataset)

- **Feature Significance:** To determine whether all features are significant in detecting CKD, one way ANOVA is performed. The feature pot had the p value of 0.189278, which is larger than significance level of 0.05. So, it was determined non-significant, and thus pot was removed from the dataframe for additional analysis.
- **Standardization:** To scale the features of dataset in the same way, standardization is performed so that the feature mean becomes 0 and standard deviation is 1. This is to ensure that each feature can take the same share of contribution to the analysis otherwise, a feature could overpower another because of its scale.
- **Feature Selection:** The Simulated Annealing [19] (SA) algorithm was applied for feature selection, and found 14 features as important to detect Chronic Kidney Disease (CKD). In figure 2 it is observed that the decreasing gap between average and maximum fitness, which means the algorithm is discovering best feature set. Algorithm 2 provides steps for feature selection using SA feature selection.

**Algorithm 2** Simulated Annealing for Feature Selection

**Input:** Initial population of features
**Output:** Best selected features
1. Initialize current_solution with a random subset of features
2. Set best_solution = current_solution 3. Set best_score = fitness_function(best_solution)
**for** iteration = 1 **to** max_iter **do**
   a. Generate a neighbor_solution by flipping a random bit in current_solution b. Calculate fitness_score of neighbor_solution c. **if** neighbor_score ¡ best_score or with probability $\exp \frac{best\_score - neighbor\_score}{temperature}$ **then**
      Update current_solution
   d. Update best_solution and best_score if needed e. Record average and max fitness values f. Cool down the temperature
**return** best_solution as the selected features

- **Splitting the dataset:** An 80:20 ratio is used to split the dataset into training and test sets, with 80% of the data used for training and 20% for testing. There are 317 samples in the training set and 80 samples in the test set.

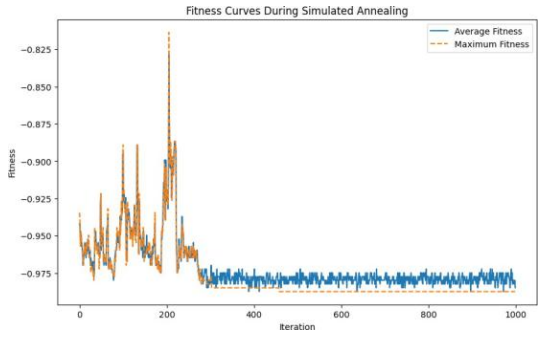

Fig. 2. Max Fitness vs Average Fitness

- **Data Balancing:** An imbalance of the training set exists as 197 samples for label 0 and 120 samples for label 1.We use SMOTE to balance the dataset to fix this.We increase both labels to 450 samples each so that the model can learn as well from both classes. So it helps prevent overfitting, and the model performs better.
- **Model Training and Evaluation:** Four models were used in this study: Support Vector Classifier (SVC), Multilayer Perceptron (MLP), Random Forest and CatBoost. Grid Search was used to find the best settings for each model. All models were regularized to prevent over fitting. Additionally, we also applied 5-Fold cross validation to check model that they are doing well across data subsets and generalize well.

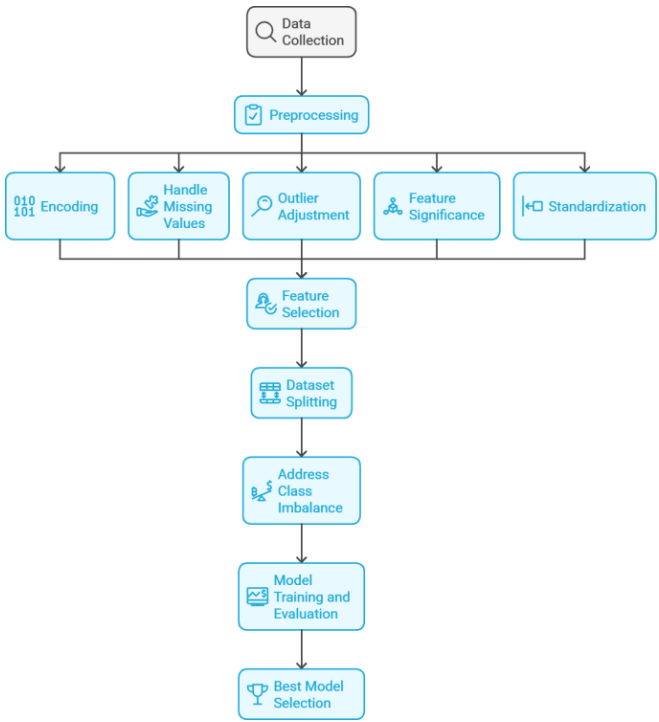

Fig. 3. Workflow Diagram

## IV. EVALUATION METRICS

The models were evaluated using key metrics: accuracy, precision, recall, F1-score, AUC-ROC, and confusion matrix. All metrics offer a view of how well the model is performing. The table II provides the formulas and descriptions for these metrics.

TABLE II
EVALUATION METRICS AND THEIR FORMULAS

| Metric | Formula | Description |
| --- | --- | --- |
| Accuracy | $\frac{TP+TN}{TP+TN+FP+FN}$ | Proportion of correct predictions. |
| Precision | $\frac{TP}{TP+FP}$ | Proportion of positive predictions that are correct. |
| Recall | $\frac{TP}{TP+FN}$ | Proportion of actual positives correctly identified. |
| F1-Score | $2 \times \frac{Precision \times Recall}{Precision+Recall}$ | Harmonic mean of precision and recall. |
| AUC-ROC | Area under the ROC curve | Measures the trade-off between sensitivity and specificity. |
| Confusion Matrix | $\begin{bmatrix} TP & FP \\ FN & TN \end{bmatrix}$ | Displays true positives, false positives, false negatives, and true negatives. |

## V. RESULTS AND DISCUSSION

Four different models were used to evaluate their performance in detection of chronic kidney disease (CKD) in this experiment. The models tested include: Logistic Regression (LR), Multilayer Perceptron (MLP, Random Forest (RF), Cat-Boost.
The training and testing results for these models are provided below to identify the most effective model for detecting CKD.

### A. Training & Validation Results

The table III shows the training accuracy and 5-fold cross-validation accuracies. Accuracy of Logistic Regression is 98.11% and 5 fold cross validation accuracy is 97.67%. The best results were obtained by MLP (Multi-layer Perceptron) with an accuracy of 99.89% and a 5 fold cross validation accuracy of 99.33%. Random Forest achieves an accuracy of 99.69% and a 5 fold cross validation accuracy of 99.22%. Finally, 99.86% accuracy is achieved by CatBoost and 99.56% is achieved by a 5-fold cross validation. These results suggest that all models generalize well, performing well on both training and validation accuracies with only small differences between the two.
Table IV demonstrates precision, recall and F1 score for both classes to show more about model performance especially for dealing with class imbalance and identifying positive and negative classes.

TABLE III
MODEL ACCURACY AND VALIDATION ACCURACY

| Metric | LR | MLP | RF | CatBoost |
| --- | --- | --- | --- | --- |
| **Accuracy** | 0.9811 | 0.9989 | 0.9969 | 0.9986 |
| **5-Fold CV Accuracy** | 0.9767 | 0.9933 | 0.9922 | 0.9956 |

The performance metrics for four models across class 0 and class 1 are summarized in Table IV, including precision, recall and F1 score. The results for logistic regression are solid, with class 0 precision of 0.9843 and class 1 precision of 0.9779. On each of these datasets, MLP obtains high scores, with a precision of 0.9990 for class 0 and 0.9978 for class 1. Random Forest and CatBoost have similar performance, both having very good precision, recall and F1 scores, mostly for class 0. Random forest is slightly outperformed by CatBoost, especially in class 0 metrics. These results show the good performance of the models in separating both classes.

In terms of computational efficiency, each model is evaluated by training time, which is reported in Table V.

TABLE IV
MODEL PERFORMANCE METRICS (CLASS 0 AND CLASS 1)

| Model | Class | Precision | Recall | F1-Score |
|---|---|---|---|---|
| Logistic Regression | 0 | 0.9843 | 0.9778 | 0.9810 |
| | 1 | 0.9779 | 0.9844 | 0.9812 |
| MLP | 0 | 0.9990 | 0.9978 | 0.9984 |
| | 1 | 0.9978 | 1.0000 | 0.9989 |
| Random Forest | 0 | 0.9975 | 0.9970 | 0.9978 |
| | 1 | 0.9963 | 0.9966 | 0.9962 |
| CatBoost | 0 | 0.9994 | 0.9992 | 0.9993 |
| | 1 | 0.9978 | 0.9992 | 0.9985 |

Training times for Logistic Regression, MLP, Random Forest, and CatBoost are shown in the table V. MLP takes the longest at 3.918 seconds, and is the slowest, followed by Logistic Regression which is the fastest and require only 0.0249 seconds. Across the models, we see differences in terms of computational efficiency, and this is shown by the moderate training time of 0.1299 seconds for Random Forest and 0.5405 seconds for CatBoost.

A learning curve is used to estimate how a model improves as it get more data and to measure its efficiency and generalization ability. Previously we analyzed training and cross-validation accuracy, we will review evaluation metrics and training time. Figure 4 shows the learning curve of all model, how validation accuracy improves over training accuracy.

TABLE V
TRAINING TIME FOR MODELS

| Model | LR | MLP | RF | CatBoost |
|---|---|---|---|---|
| Training Time (seconds) | 0.0249 | 3.9180 | 0.1299 | 0.5405 |

It is shown in the figure below that model validation accuracy improves over time, with validation accuracy increasing fold by fold. Overall, the CatBoost model's validation accuracy is somewhat more consistent compared to the other models, demonstrating a steadier improvement throughout the training process.

### B. Testing Results

The testing results of all trained model to detect chronic kidney disease are presented in table VI. It is find that Logistic Regression achieves 96.25% accuracy and AUC of 0.9663 which is fairly good on classification. Also, Random Forest has an accuracy of 97.50% and an AUC of 0.9980, which is strong in order to differentiate classes. The best results are obtained by CatBoost with 98.75% accuracy and 0.9993 AUC, both in terms of the prediction accuracy and class distinguishing ability.

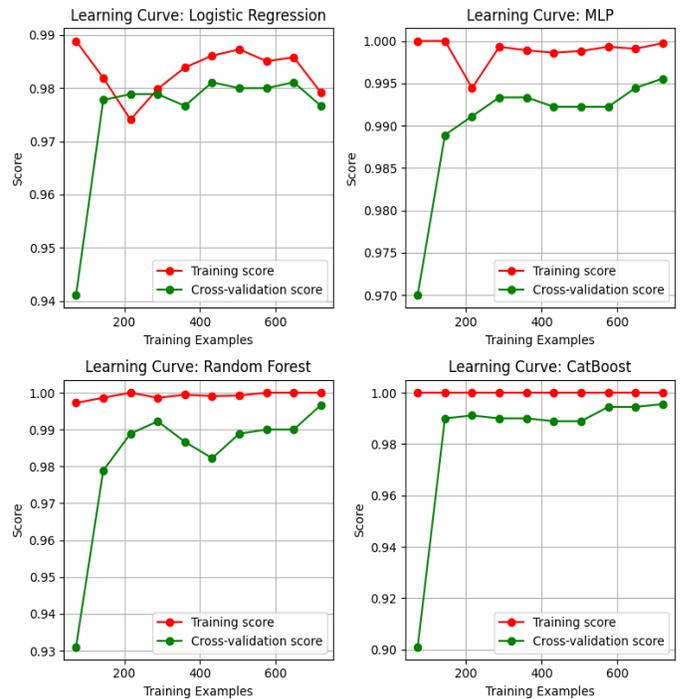

Fig. 4. Learning Curve

In Table VII, testing evaluation metrics are presented to give a more detailed view of the performance of each model compared to accuracy and AUC.

TABLE VI
MODEL ACCURACY AND AUC SCORES

| Metric | LR | MLP | RF | CatBoost |
|---|---|---|---|---|
| Accuracy | 96.25% | 96.25% | 97.50% | 98.75% |
| AUC Score | 0.9663 | 0.9778 | 0.9980 | 0.9993 |

The table VII presents precision, recall, and F1-score metrics for class 0 and class 1 across four models. Logistic Regression and MLP are both found to be perfect precision for class 0, and lower precision for class 1, and perfect recall for class 1. Random Forest maintains good precision and recall and balanced performance between the two classes. CatBoost achieves perfect precision for class 0 and perfect recall for class 1 making it the best model in this comparison. Figure 5 includes the confusion matrix, which gives a breakdown of model predictions to show how many of each were true positives, false positives, true negatives, and false negatives for each model.

As shown in figure 5, CatBoost performed the best with no False Negatives and only one False Positive, which means a better performance with little misclassifications. Both Logistic

TABLE VII
MODEL PERFORMANCE METRICS (CLASS 0 AND 1)

| Model | Class | Precision | Recall | F1-Score |
|---|---|---|---|---|
| Logistic Regression | 0 | 1.0000 | 0.9400 | 0.9691 |
|  | 1 | 0.9091 | 1.0000 | 0.9524 |
| MLP | 0 | 1.0000 | 0.9400 | 0.9691 |
|  | 1 | 0.9091 | 1.0000 | 0.9524 |
| Random Forest | 0 | 0.9800 | 0.9800 | 0.9800 |
|  | 1 | 0.9667 | 0.9667 | 0.9667 |
| CatBoost | 0 | **1.0000** | 0.9800 | 0.9899 |
|  | 1 | 0.9677 | **1.0000** | 0.9836 |

Regression and MLP models had zero False Negatives, but each had three False Positives. Random Forest had 1 False Positive and 1 False Negative.

Figure 1 includes the ROC curve to provide a deeper comparison of each model's ability to distinguish between classes.

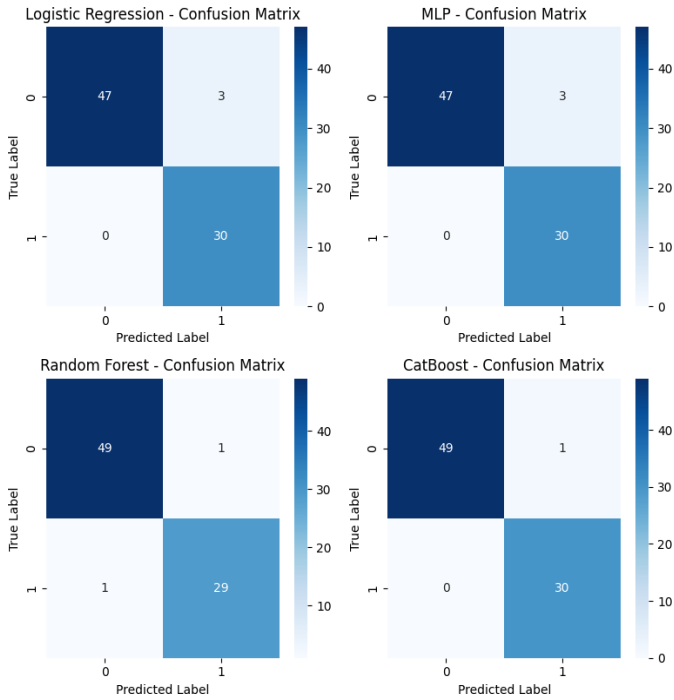

Fig. 5. Confusion Matrix

Figure 6 shows the ROC curve and it can be seen that, as nearly a flat line in the top left corner, CatBoost outperforms all other models. Also, Random Forest perform very well, following CatBoost closely. On the other hand, the curve for Logistic Regression and MLP is distorted, it means these two models struggle bit to distinguishing between the classes.

## VI. KAPPA ASSESSMENT

Although Logistic Regression, Multi-layer Perceptron, and Random Forest exhibit high accuracy, their relatively lower Cohen's Kappa scores suggest potential issues with class imbalance or reliance on the majority class. In contrast, CatBoost shows a Cohen's Kappa value close to its accuracy, indicating a stronger alignment with the true labels and a more balanced, reliable performance across all classes that is shown from table VIII.

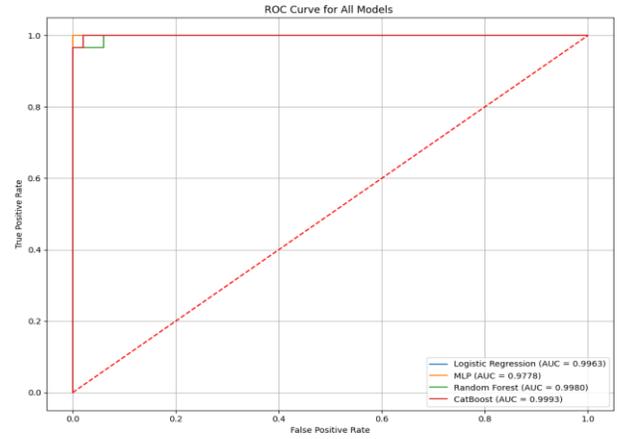

Fig. 6. ROC Curve

TABLE VIII
COHEN'S KAPPA SCORES

| Model | LR | MLP | RF | CatBoost |
|---|---|---|---|---|
| Cohen's Kappa | 0.9216 | 0.9216 | 0.9467 | 0.9735 |

## VII. PROPOSED APPROACH

After reviewing the training accuracy, testing accuracy, validation accuracy, evaluation metrics for train and test sets, training time, confusion matrix, ROC curve and kappa score, it is noticeable that CatBoost is the proposed model for CKD detection in a realiable and effective way. The analysis displays that CatBoost surpass other models, offering reliable and better performance with consistent metrics and a nearly perfect ROC curve. The parameters used for this experiment are summarized below in Table IX

TABLE IX
PARAMETERS USED FOR CATBOOST IN CKD DETECTION

| border_count | depth | iterations | l2_leaf_reg | learning_rate | verbose | regularization |
|---|---|---|---|---|---|---|
| 32 | 8 | 200 | 3 | 0.01 | 0 | Enabled |

## VIII. SHAP FOR MODEL DECISION MAKING INTERPRETATION

To understand the model's decision-making process for Chronic Kidney Disease (CKD) detection, we apply SHAP to the test set to identify which features influence the model's predictions, as shown in Figure 7.

The analysis of SHAP plot shows that the specific gravity (sg), serum creatinine (sc), albumin (al), hemoglobin (hemo), and diabetes mellitus (dm) are highly influential to predict CKD. CKD is strongly associated with high sg and low albumin (red on the left), and high hemoglobin (red on the

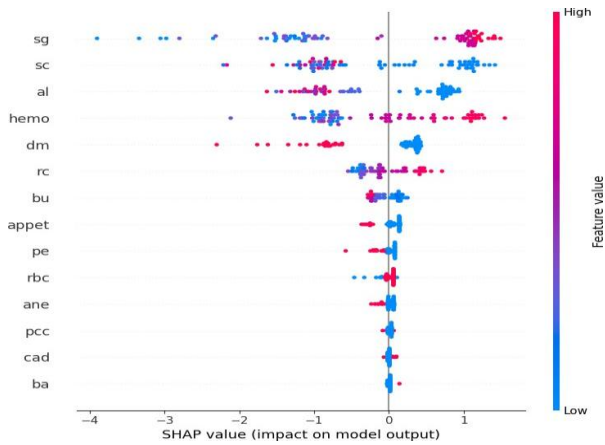

Fig. 7. SHAP Summary Plot

right).Also, high values (red) on the left indicate that the presence of diabetes decreases the likelihood of CKD in the model, possibly due to dataset specific interactions or confounding factors for diabetes mellitus (dm). These observations are consistent with clinical insights, but the surprising association with diabetes mellitus may warrant further work to understand the effect of diabetes mellitus on the predictions of the model, which would improve the model's interpretability and reliability.

## IX. COMPARATIVE ANALYSIS

From Table X it is shown that our study outperforms most existing methods compared to previous studies on CKD detection using the same dataset. It was achieved using nature inspired feature selection and an optimized CatBoost algorithm. Finally, these findings underscore the impact of our approach to improve CKD detection accuracy and address other CKD prediction challenges.

TABLE X
COMPARATIVE STUDY OF CKD PREDICTION MODELS USING UCI CKD DATASET

| Study | Dataset | Accuracy (%) |
|---|---|---|
| Shahid et al. [8] | UCI CKD Dataset | 98.47 (AdaBoost) |
| Gangani et al. [9] | | 97.5 (XGBoost) |
| Saurabh Pal [10] | | 93 (Random Forest) |
| Ariful et al. [11] | | 98.3 (Predictive Modeling) |
| Nayeem et al. [12] | | 98.61 (ANN) |
| Pal [13] | | 97.23 (Bagging + DT) |
| Hamida et al. [14] | | 85.5 (J48) |
| **Our Study** | | **98.75 (Tuned CatBoost)** |

## X. CONCLUSION AND FUTURE WORK

This study brings an improved approach of CKD detection by machine learning using the Catboost model with the accuracy of 98.75% and AUC of 0.9993 which outperforms the other models. Simulated Annealing and Cuckoo Search are further applied to feature selection, and outlier adjustment for the model to make it robust. To achieve transparency, explainable AI (XAI) technique is integrated in this study so as to demonstrate the feasibility of XAI methods in CKD detection and clinical decision making. In future work the model should be extended to larger and more diverse datasets or real time applications or website integration. Finally, the model will be explored with deep learning models and be collaborating with a number of healthcare professionals to incorporate wider datasets in order to make the model robust and applicable in real world clinical settings.